\documentclass[10pt,twocolumn,letterpaper]{article}

\pdfoutput=1

\usepackage{cvpr}
\usepackage{times}
\usepackage{graphicx}
\usepackage{cite}
\usepackage{fixltx2e}   % Fix fullpage figure ordering
\usepackage{amsmath,amssymb} % define this before the line numbering.
\usepackage{color}
\usepackage{url}
\usepackage{tikz}
\usepackage{algorithm}
\usepackage{algorithmicx}
\usepackage{algpseudocode}
\usepackage{booktabs}
\usepackage{paralist}

\newcommand{\vitto}[1]{\textcolor{red}{[VF: #1]}}
\newcommand{\rahul}[1]{\textcolor{blue}{[RS- #1]}}
\newcommand{\luca}[1]{\textcolor{orange}{[LUCA- #1]}}
\newcommand{\ricco}[1]{\textcolor{purple}{[RICCO- #1]}}
\newcommand{\google}[1]{\textcolor{red}{[GOOGLE- #1]}}
\renewcommand{\rahul}[1]{}
\renewcommand{\luca}[1]{}
\renewcommand{\ricco}[1]{}
\renewcommand{\google}[1]{}
\renewcommand{\vitto}[1]{}
\DeclareMathAlphabet{\mathpzc}{T1}{pzc}{m}{n}
\newlength{\halfwidth}
\newlength{\fullwidth}
\usetikzlibrary{shapes}
\pgfdeclarelayer{background}
\pgfsetlayers{background,main}
\newlength{\tikzimgheight}
\newlength{\tikzimgwidth}

\usepackage{ifthen}
\usetikzlibrary{calc}

\algrenewcommand{\algorithmiccomment}[1]{\hfill \textcolor{blue}{[#1]}}

\usepackage[pagebackref=true,breaklinks=true,letterpaper=true,colorlinks,bookmarks=false]{hyperref}

\cvprfinalcopy % *** Uncomment this line for the final submission

\def\cvprPaperID{1013} % *** Enter the CVPR Paper ID here

\ifcvprfinal\fi
\begin{document}

%%%%%%%%% TITLE
\title{Articulated Motion Discovery using Pairs of Trajectories}

\author{Luca Del Pero\textsuperscript{1} \hspace{1.22cm} Susanna Ricco\textsuperscript{2} \hspace{1.22cm} Rahul Sukthankar\textsuperscript{2} \hspace{1.22cm} Vittorio Ferrari\textsuperscript{1}\\
{\tt\small \hspace{-0.2cm} ldelper@inf.ed.ac.uk \hspace{0.3cm} ricco@google.com \hspace{0.5cm} sukthankar@google.com  \hspace{0.1cm} ferrari@inf.ed.ac.uk}\\
\hspace{-0.6cm} \textsuperscript{1}University of Edinburgh \hspace{0.9cm}\textsuperscript{2}Google Research
}

\maketitle
%\thispagestyle{empty}

%%%%%%%%% ABSTRACT
\begin{abstract}
We propose an unsupervised approach for discovering
characteristic motion patterns in videos of highly articulated
objects performing natural, unscripted behaviors, such as tigers in the wild.
We discover consistent patterns in a bottom-up manner by analyzing
the relative displacements of large numbers of ordered trajectory
pairs through time, such that each trajectory is attached to a
different moving part on the object. The pairs of trajectories
descriptor relies entirely on motion
and is more
discriminative than state-of-the-art features that employ single
trajectories.
Our method generates temporal video intervals, each automatically trimmed to one instance of
the discovered behavior, and clusters them by type (e.g., running, turning head, drinking water).
We present experiments on two datasets: dogs from YouTube-Objects
and a new dataset of National Geographic tiger videos.
Results confirm that our proposed descriptor outperforms existing appearance-
and trajectory-based descriptors (e.g., HOG and DTFs) on both datasets 
and enables us to segment unconstrained animal video into intervals containing
single behaviors.
\end{abstract}

%%%%%%%%% BODY TEXT
\vspace{-10pt}
\section{Introduction}
Internet videos provide a wealth of data that could be used to learn the
appearance or expected behaviors of many object classes. However,
traditional supervised learning techniques used on still images~\cite{Cootes1998ECCV,Felzenszwalb03pictorialstructures,BourdevMalikICCV09}
do not easily transfer due to the prohibitive cost of generating ground-truth annotations in videos.
In order to realize the full potential of this vast resource, we must instead rely
on methods that require as little human supervision as possible.

\begin{figure}[t]
\begin{center}
\includegraphics[width=0.32\textwidth]{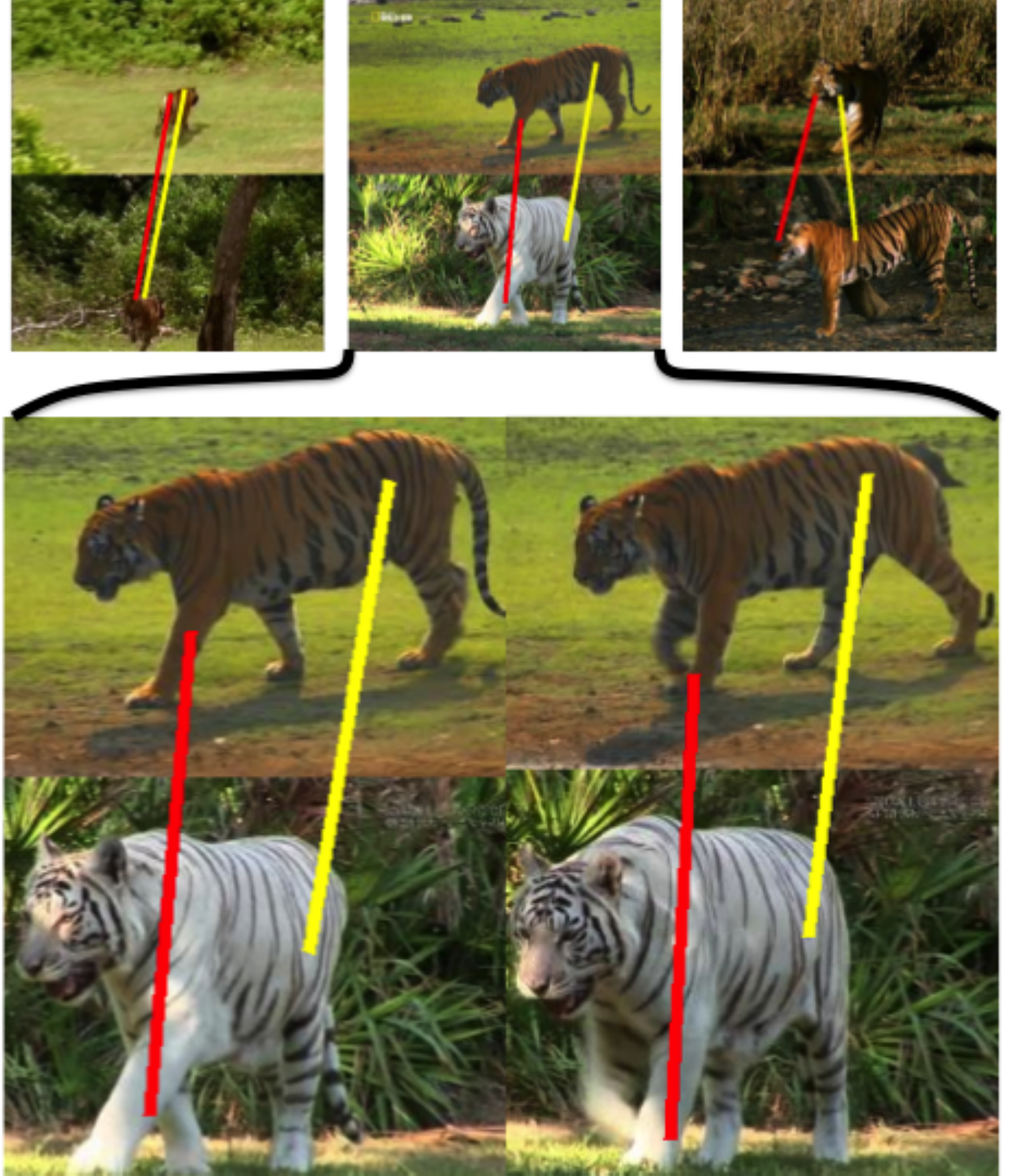}
\end{center}
\caption{\small
  Examples of articulated motion pattern clusters discovered using pairs
  of trajectories (PoTs).  These clusters capture tigers running,
  walking, and turning their heads, respectively.  Inset
  shows detail of one PoT within the walking cluster.
  Yellow lines connect the first trajectories (on the tigers' body);
  red lines connect the second (on moving extremities).}
\label{fig:teaser}
\end{figure}

We propose a bottom-up method for discovering the characteristic motion
patterns of an articulated object class \emph{in the wild}.  Unlike the
majority of action recognition datasets, in which human actors perform 
scripted actions~\cite{MSRActions, UTInteraction, KTH, WeizmannActions}, and/or clips are 
trimmed to contain a single action~\cite{HMDB, UCF101}, our videos are
unstructured, such as animals performing unscripted behaviors.  The only assumption we make is that each video contain at least one
instance of the object class.  We leverage that the object is engaged in some (unknown) behaviors, and that
such behaviors exhibit observable consistency, which we term characteristic
\emph{motion patterns}.

Our method does not require knowledge of the number or
types of behaviors, nor that instances of different behaviors be temporally
segmented within a video.
The output of our method is a set of video intervals, clustered according to
the observed characteristic motion patterns.  Each interval contains one
temporally segmented instance of the pattern.  Fig.~\ref{fig:teaser} shows some
behaviors automatically discovered 
in tiger videos, such
as walking, turning head, and running.

We identify consistency between observed motion patterns by analyzing the relative
displacement of large numbers of ordered trajectory pairs (PoTs). The first trajectory
in the pair defines a reference frame in which the motion of the second
trajectory is measured.  We preferentially sample trajectory pairs across
joints, resulting in features particularly well-suited to representing
fine-grained behaviors of complex, articulated objects. This has greater
discriminative power than state-of-the-art features defined using single
trajectories in isolation~\cite{Wang_2011_CVPR,wang_ICCV_2013}.

\begin{figure*}[t]
\begin{center}
%\includegraphics[scale =0.17]{images/tigersclean.pdf}
%
%\vspace{6pt}
%
%\includegraphics[scale =0.28]{images/fig2temporal.pdf}
\includegraphics[scale =0.48]{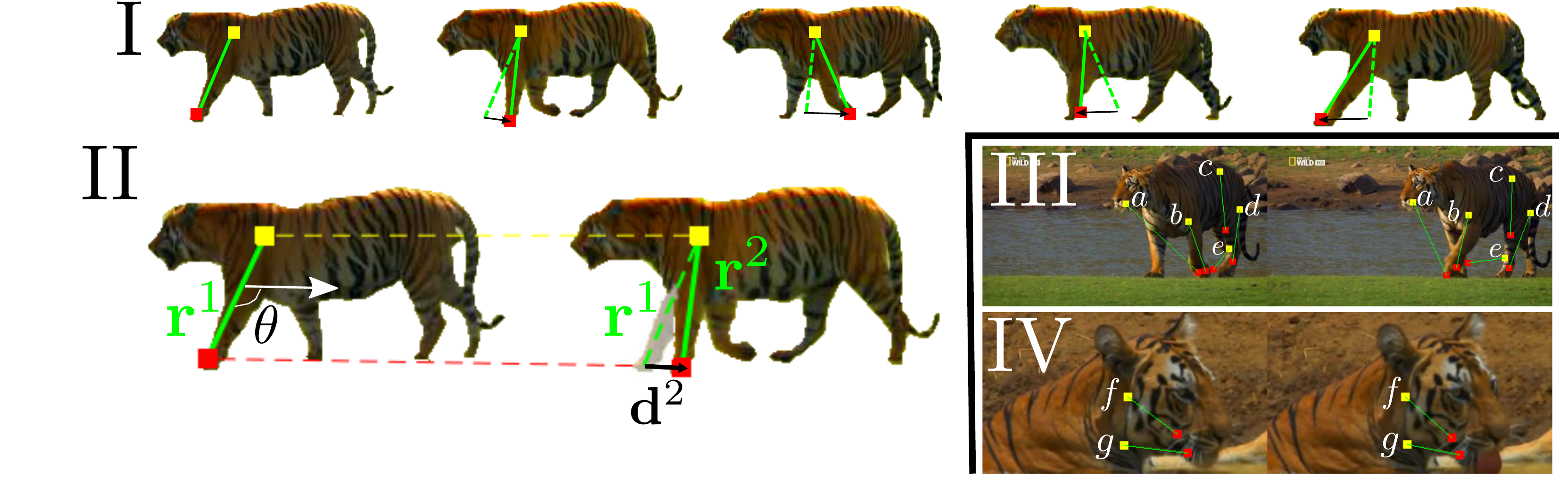}
\end{center}
 \caption{\small{
\vitto{size of $\theta$}
Modeling articulated motion with PoTs.
Two trajectories in a PoT are ordered  based on their deviation from the median
velocity of the object: the anchor (yellow) deviates less than the swing (red).
In I, the displacement of the swing relative to the anchor follows the swinging
motion of the paw with respect to the shoulder. While both move forward as the
tiger walks, the paw is actually moving backwards in a coordinate system
centered at the shoulder. This back-and-forth motion is captured by the
relative displacement vectors of the pair (in black) but missed when individual
trajectories are used alone.  The PoT descriptor is constructed from the angle
$\theta$ and the black vectors $\mathbf{d^k}$, shown in II.  The two
trajectories in a PoT are selected such that they track object parts that 
move differently. A few selected PoTs are shown in III and IV. Legs move
differently than the head (a), hip (c), knees (b,d), or other legs (e). In IV,
the head rotates relative to the neck, resulting in different PoTs (f,g).  Our
method selects these PoTs without requiring prior knowledge of the object
topology.
 } } \vspace{-3pt}\label{fig:PoTIntro} \end{figure*}

Although we often refer to PoTs using semantic labels for the location of their
component trajectories (eye, shoulder, hip, etc.), these are used only for
convenience.  PoTs do not require semantic understanding or any part-based or
skeletal model of the object, nor are they specific to an object class.
Furthermore, the collection of PoTs is more expressive than a simple star-like
model in which the motion of point trajectories are measured relative to the
center of mass of the object. 
For example, we find the ``walking" cluster (Fig.~\ref{fig:qualitative}) 
based on PoTs formed by various combinations
of  head-paw (Fig.~\ref{fig:PoTIntro} III, a), hip-knee (c), knee-paw (b,d),
or even paw-paw (e) trajectories. 
%For example, we find the walking cluster based on
%PoTs formed by various combinations of shoulder-paw, head-paw, hip-paw, or even
%paw-paw trajectories (see
%Figs.~\ref{fig:PoTexamples}~and~\ref{fig:qualitative}).

In contrast to other popular descriptors~\cite{Jain2013, Wang_2011_CVPR,
wang_ICCV_2013}, PoTs are appearance-free. They are defined solely by motion
and so are robust to appearance variations within the object class. In cases
where appearance proves beneficial for discriminating between behaviors of
interest, it is easy to combine PoTs with standard appearance features.

In summary, our main contributions are:
(1) a new feature based on ordered pairs of trajectories that captures the
intricate motion of articulated objects (Sec.~\ref{sec:PoTs});
(2) a method for unsupervised discovery of behaviors from unconstrained videos of an object class (Sec.~\ref{sec:discovery});
(3) a method for identifying periodic motion in video, which we use to segment videos into intervals containing single behaviors (Sec.~\ref{sec:partitioning}); and
(4) annotations for 80,000 frames from nature documentaries about tigers and
20,000 frames from YouTube videos of dogs (Sec.~\ref{sec:experiments}),
available on our website~\cite{delpero15cvpr-potswebpage}.

% The extracted motion primitives could be used as a building block for
% higher-level recognition tasks. For example, complex motions
% like walking, running, and jumping,
% could all be modeled as sequences of various motion primitives involving the
% leg. The \textbf{temporal correspondences} could allow sequences containing these
% motions to be temporally aligned. The recovered \textbf{spatial correspondences} 
% could be used for unsupervised decomposition of an object into independently moving
% parts. In turn, this would allow unsupervised learning of the appearance of the
% parts and their valid spatial configurations.

\section{Related work}
Motion is a fundamental cue for many applications in video analysis and so has
been widely studied, particularly within the context of action
recognition~\cite{Turaga2008,Weinland2010}. However, action recognition is
traditionally formulated as a supervised classification
problem~\cite{HMDB51,UCF101}.
Work on unsupervised motion analysis has largely
been restricted to the problem of dynamic scene analysis~\cite{kuettel10cvpr,
HospedalesICCV09, Mahadevan2010, Wang2009PAMI, Hu2006PAMI, Zhao2011}.  These
works typically consider a fixed scene observed at a distance from a static
camera; the goal is to model the behavior of agents (typically pedestrians
and vehicles) and to detect anomalous events. Features typically consist of
optical flow at each pixel~\cite{HospedalesICCV09,kuettel10cvpr,Wang2009PAMI}
or single trajectories corresponding to tracked
objects~\cite{Hu2006PAMI,Zhao2011}.

\luca{I changed a bit this paragraph, please check}
To our knowledge, only Yang \etal~\cite{Yang_PAMI_2013} considered the task of
unsupervised motion pattern discovery, although from manually trimmed videos.
Their method models human actions in terms of motion primitives discovered by 
clustering localized optical flow vectors, normalized with respect to the dominant translation of the object.
%Their preprocessing steps for removing camera or
%significant object motion result in features that are normalized with respect
%to the dominant translation of the object. 
In contrast, our pairwise features capture complex relationships between the motion of two different object parts.
Furthermore, we describe motion at a more informative temporal scale by using
multiframe trajectories instead of two-frame optical flow.
We compare experimentally to~\cite{Yang_PAMI_2013} on the KTH dataset~\cite{KTH} in Sec.~\ref{sec:evalpots}.
% While they share our goal of finding recurring motion
% patterns by observing a class of articulated objects, their features
% are unlikely to generalize to the variety of behaviors and viewing conditions found in
% unstructured video.

Although many approaches do not easily transfer from the supervised to the
unsupervised domain, one major breakthrough from the action recognition
literature that does is the concept of {\em dense trajectories}.
The idea of generating trajectories for each object from large numbers of KLT interest
points in order to model its articulation was simultaneously proposed by
Matikainen~\etal~\cite{Matikainen2009} and Messing~\etal~\cite{Messing2009} for
action recognition.
These ideas were extended and refined in the work on
tracklets~\cite{raptis10eccv} and dense trajectory features
(DTFs)~\cite{Wang_2011_CVPR,wang_ICCV_2013}. DTFs currently
provide state-of-the-art performance on video action recognition~\cite{THUMOS2014}.

\luca{Please also check this paragraph, added stuff on Narayan and compressed the rest}
In contrast to our work, most trajectory-based methods treat each trajectory in 
isolation~\cite{Wang_2011_CVPR,wang_ICCV_2013, Messing2009, Matikainen2009,raptis10eccv},
with two notable exceptions~\cite{Jiang2012,narayan14cvpr}.
Jiang \etal~\cite{Jiang2012} assign individual trajectories to a single codeword from a predefined
codebook (as in DTF works~\cite{Wang_2011_CVPR,wang_ICCV_2013}).
However, the codewords from a pair of trajectories are combined
into a `codeword pair' augmented by coarse information about the relative motion
and average location of the two trajectories. Yet, this pairwise analysis is
cursory: the selection of codewords is unchanged from the single-trajectory
case, and the descriptor thus lacks the fine-grained information about the
relative motion of the trajectories that our proposed PoTs provide.
Narayan \etal~\cite{narayan14cvpr} model Granger causality between trajectory codewords. 
Their global descriptor only captures pairwise statistics of codewords 
over a fixed-length temporal interval. In contrast, a PoT groups two trajectories into a single 
local feature, with a descriptor encoding their spatiotemporal arrangement. Hence, PoTs can be used to 
find point correspondences between videos (Fig.~\ref{fig:qualitative}).

The few remaining methods that propose pairwise representations employ them in
a very different context.
Leordeanu~\etal~\cite{Leordeanu2007} learned object classes from still images
by matching pairs of contour points from one image to pairs in another.
Yang~\etal~\cite{Yang2010} computed statistics between local feature pairs
for food recognition in images.
Matikainen~\etal~\cite{Matikainen2010} used spatial and temporal features
computed over pairs of sparse KLT trajectories to construct a two-level
codebook for action classification.
Dynamic-poselets~\cite{Wang2014} requires detailed manual annotations of human
skeletal structure on training data 
to define a descriptor for pairs of connected joints. 
\luca{Added the sentence on Raptis, and compressed the paragraph a bit}
Raptis \etal~\cite{raptis12cvpr} consider pairwise interactions between
clusters of trajectories, but their method also
requires detailed manual annotation for each action.
None of these approaches is suitable for 
unsupervised articulated motion discovery.

A few recent works exploit video as a source of training data for object class detectors~\cite{prest12cvpr,Tang2013}. They separate object instances from their background based on motion, thus reducing the need for manual bounding-box annotation. However, their use of video stops at segmentation. They make no attempt at modeling articulated motion or finding common motion patterns across videos.
Ramanan et al.~\cite{ramanan06pami} build a 2D part-based model of an animal from one video. The model is a pictorial structure based on a 2D kinematic chain of coarse rectangular segments.
Their method operates strictly on individual videos and therefore cannot find motion patterns characteristic for a class. It is tested on just three simple videos containing only the animal from a single, unchanging viewpoint.

\vspace{-4pt}
\section{Pairs of Trajectories (PoTs)} 
\label{sec:PoTs} 

We represent articulated object motion using a collection of
\emph{automatically selected} ordered pairs of trajectories (PoTs),
tracked over $n$ frames. Only two trajectories following parts 
of the object moving relatively to each other are selected as a PoT, as
these are the pairs that move in a consistent and distinctive manner
across different instances of a specific motion pattern.
For example, the motion of a pair connecting a tiger's knee to
its paw consistently recurs across videos of walking tigers (Figs.~\ref{fig:teaser} and~\ref{fig:qualitative}).
By contrast, a pair connecting two points on the chest (a
rather rigid structure) may be insufficiently distinctive,
while one connecting the tip of the tail to the nose may lack consistency.
Note also that a trajectory may simultaneously contribute to multiple
PoTs (\eg, a trajectory on the front paw may form pairs with trajectories from
the shoulder, hip, and nose).

Fig.~\ref{fig:PoTIntro} (III-IV) shows a few examples of PoTs selected from two tiger videos. We define PoTs 
%and their motion descriptor 
in Sec.~\ref{sec:definition}, while we explain how to select PoTs from real videos in Sec.~\ref{sec:extraction}.

\begin{figure*}[t]
\begin{center}
\setlength{\tabcolsep}{0.8pt}
\begin{tabular}{c c c c c c}
\footnotesize{\textbf{input foreground}} & \footnotesize{\textbf{deviation from}} & \footnotesize{\textbf{extracted PoTs}} & \footnotesize{\textbf{anchors and swings}}  & \footnotesize{\textbf{extracted PoTs}} & \footnotesize{\textbf{anchors and swings}} \\[-3pt]
\footnotesize{\textbf{trajectories}} & \footnotesize{\textbf{median velocity}} & \scriptsize{$\mathbf{(\theta_{P}=0.01)}$} & \scriptsize{$\mathbf{(\theta_{P}=0.01)}$}  & \scriptsize{$\mathbf{(\theta_{P}=0.15)}$} & \scriptsize{$\mathbf{(\theta_{P}=0.15)}$} \\
\includegraphics[scale =0.19]{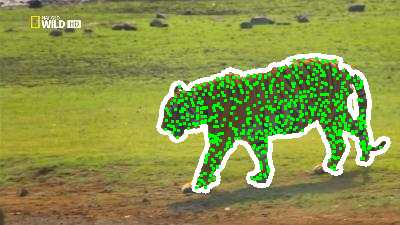}  &
\includegraphics[scale =0.19]{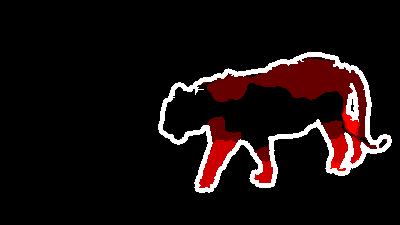}  &
\includegraphics[scale =0.19]{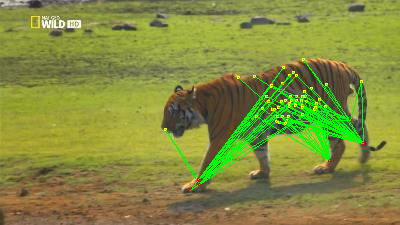}  &
\includegraphics[scale =0.19]{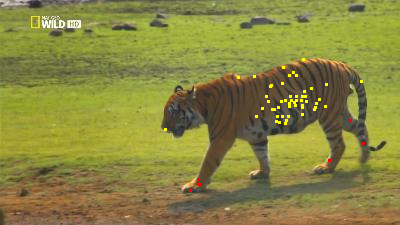}  &
\includegraphics[scale =0.19]{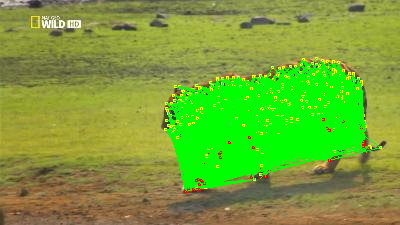}  &
\includegraphics[scale=0.19]{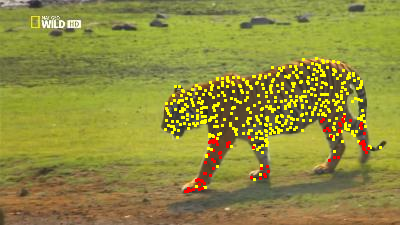} \\
\includegraphics[scale =0.19]{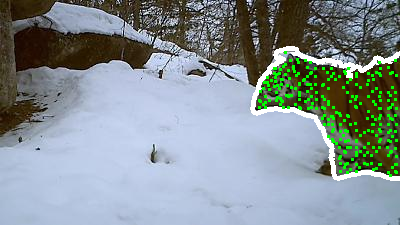}  &
\includegraphics[scale =0.19]{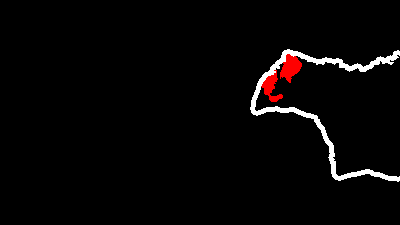}  &
\includegraphics[scale =0.19]{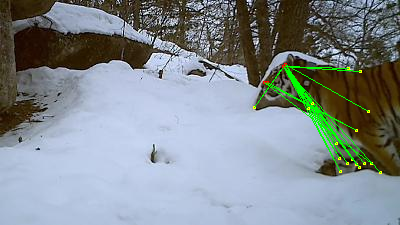}  &
\includegraphics[scale =0.19]{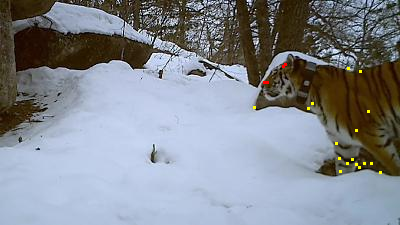}  &
\includegraphics[scale =0.19]{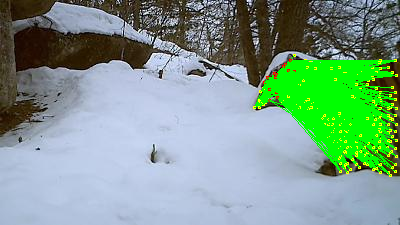}  &
\includegraphics[scale=0.19]{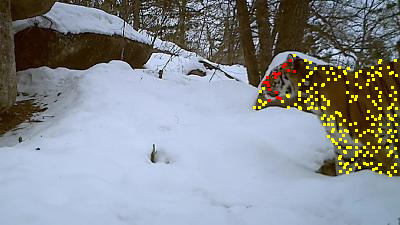} \\
\small{(a)} & \small{(b)} & \small{(c)} & \small{(d)} & \small{(e)} & \small{(f)} \\
\end{tabular}
\end{center}
 \caption{\small{
PoT selection on two different examples: a tiger walking (top) and one turning
its head (bottom). We construct PoT candidates from the trajectories on the
foreground mask (a), using all possible pairs. We prefer candidates where the
anchor is closer to the median foreground velocity, denoted by dark areas in (b), while the
swing follows a different motion (bright areas). We keep the highest
$\theta_{P} \%$ ranking candidates according to this criterion. We show the
selected PoTs for two different values of $\theta_{P}$ (c,e).
Too strict a $\theta_{P}$ ignores many interesting PoTs, like those involving trajectories
on the neck in the top row (c).
We also show the trajectories used as anchors (yellow)  and swings (red) without the lines connecting them 
(d,f). Imagine connecting any anchor with any
swing: in most cases, the two follow different, independently moving parts of
the object, which is the key requirement of a PoT. 
We use $\theta_{P} = 0.15$ in our experiments (e,f).
}
}
\label{fig:PoTextraction} \end{figure*}

\subsection{PoT definition}
\label{sec:definition}
\paragraph{Anchors and swings.} 
The first trajectory in each PoT (the \emph{anchor})
defines a local coordinate frame, in which the motion of the second
(\emph{swing}) is measured.
We select as anchor the trajectory whose velocity is closer to the median velocity of 
pixels detected to be part of the foreground (Sec.~\ref{sec:extraction}), aggregated over the length of the PoT (this approximates the median velocity of the whole object).
This criterion generates a stable ordering, repeatable across the broad range
of videos we examine.
For example, the trajectories on the legs in Fig.~\ref{fig:PoTIntro} (I-II-III)
are consistently chosen as swings
while those on the torso are selected as anchors.

%This is particularly suitable for articulated
%objects, whose non-rigid parts move relatively to each other. 
%Consider for example the swinging motion 
%of the paw  relatively to the shoulder in Fig.~\ref{fig:PoTIntro}.

\vspace{-10pt}
\paragraph{Displacement vectors.}
%Let $a=(a_x^1,a_y^1,\ldots,a_x^n,a_y^n)$ and $s=(s_x^1,s_y^1,\ldots,s_x^n,s_y^n)$ 
%denote the $(x,y)$ coordinates of the anchor and swing, respectively. 
%In each frame $f_k$, we compute the vector from
%anchor to swing $\mathbf{r^k} = (s_x^k - a_x^k, \; s_y^k - a_y^k)$ (green lines in Fig.~\ref{fig:PoTIntro}).
In each frame $f_k$, we compute the vector $\mathbf{r^{k}}$ from
anchor to swing (green lines in Fig.~\ref{fig:PoTIntro}).
Starting from the second frame, a displacement vector $\mathbf{d^{k}}$ 
is computed by subtracting the vector $\mathbf{r^{k-1}}$
of the previous frame (dashed green) from the current $\mathbf{r^{k}}$ (solid green).
$\mathbf{d^k}$ captures the motion of
the swing relative to the anchor by canceling out the motion of the latter. 
Naively employing the green vectors $\mathbf{r^k}$ as raw features
does not capture relative motion as effectively because
the variation in $\mathbf{r^k}$ through time is dominated by the
spatial arrangement of anchor and swing rather than by the change in relative
position between frames.
This can be intuitively appreciated by comparing the magnitudes of the green
and black vectors in Fig.~\ref{fig:PoTIntro}.

\vspace{-10pt}
\paragraph{PoT descriptor.}
The PoT descriptor $P$ consists of two parts: 1) the initial position of the swing
relative to the anchor and 2) the sequence of normalized displacement
vectors through time:
\begin{equation}
P=\left(
	\theta,
	\frac{\mathbf{d^2}}{D}, \ldots, \frac{\mathbf{d^n}}{D}
  \right),
\end{equation} 
where $\theta$ is the angle from anchor to swing in the first frame
and the normalization factor is the total displacement $D=\sum_{k=2}^{n}||\mathbf{d^{k}}||$.
The DTF descriptor~\cite{Wang_2011_CVPR} employs a similar normalization.
Note also that the first term in $P$ records only the angle (and not
the magnitude) between anchor and swing; this retains scale invariance and
enables matching PoTs from objects of different size.
The dimensionality of $P$ is $2\cdot (n-1)+1$; in all of our experiments, we set $n = 10$.

\subsection{PoT selection}
\label{sec:extraction}

We explain here how to select PoTs from a set of input trajectories output by
a dense point tracker~\cite{wang_ICCV_2013}. We start with a summary of the process and give more details later.

First, we use a recent method for foreground segmentation~\cite{papazoglou13iccv}
to remove trajectories on the background.
Then, for each frame $f$ we build the set $\mathcal{P}_{f}$ of PoTs starting at that frame.
For computational efficiency, we directly set $\mathcal{P}_{f}=\emptyset$ for any frame unlikely to contain articulated motion. 
Otherwise, we form candidate PoTs from all pairs of 
foreground trajectories $\{t_i,t_j\}$ extending for at least $n$ frames after $f$.
Finally, we retain in $\mathcal{P}_{f}$ the candidates that are most likely to be on object parts moving relative to each other.

\vspace{-10pt}
\paragraph{Foreground segmentation.}
State-of-the-art point trajectories already attempt to limit trajectories to
foreground objects~\cite{wang_ICCV_2013}, but often fail on the wide range of videos we use.
We instead use a recent method~\cite{papazoglou13iccv} for foreground segmentation in unconstrained 
video. The resulting \emph{foreground mask} permits reliable detection
of articulated objects even under significant motion and against unconstrained
backgrounds.
Our method is robust to errors in the foreground mask because they only affect a small 
fraction of the PoT collection (Sec.~\ref{sec:evalclustering}).

In addition to removing trajectories on the background,
we also use this foreground mask to estimate the median velocity
of the object, computed as the median optical flow displacement over all pixels in the mask.

\vspace{-10pt}
\paragraph{Pruning frames without articulated motion.}
A frame is unlikely to contain articulated motion (hence PoTs) if the optical
flow displacement of foreground pixels is uniform.
This happens when the entire scene is static,
or the object moves with respect to the camera but the motion is not articulated. 
We define
$s(f) = \frac{1}{n}\sum_{i=f}^{f+n-1}\sigma_{i}$,
where $\sigma_{i}$ is the standard deviation in the optical flow displacement over
the foreground pixels at frame $i$ normalized by the mean, and $n$ the length of the PoT.
We set $\mathcal{P}_{f}=\emptyset$ for all frames where
$s(f) < \theta_{F}$, pruning frames without promising candidate pairs.
We set $\theta_{F}=0.1$ using $16$ cat videos
in which we manually labeled frames without articulated motion.
$\theta_{F}=0.1$ achieves a precision of $0.95$ and a recall of $0.75$.

\vspace{-10pt}
\paragraph{PoT candidates and selection.}
The candidate PoTs for an unpruned frame $f$ are all ordered pairs of trajectories $\{t_i,t_j\}$ 
that exist in $f$ and in the following $n-1$ frames and lie on the foreground mask. These trajectories are shown in Fig.~\ref{fig:PoTextraction}(a).
We score a candidate pair $\{t_i,t_j\}$ using
\begin{equation}
  \mathrm{S}(\{t_i=a, t_j=s\}) = \sum_{k=f}^{f+n-1}||v_{s}^{k} - v_{m}^{k}|| - ||v_{a}^{k} - v_{m}
^{k}||~~,
\label{eq:score_function}
\end{equation}
where $v_{m}^{k}$ is the median velocity at frame $k$, and
$v_{s}^{k}$ and $v_{a}^{k}$ the velocities of 
the swing and anchor, respectively.
The first term favors pairs with a large deviation between
swing and median velocity, while the second term favors pairs where the velocity
of the anchor is close to the median.
As seen in Fig.~\ref{fig:PoTextraction}, this generates a stable PoT ordering where anchors and swings fall on the core and extremities of the animal, respectively. However, note that the velocity of the anchors can vary; anchors along the tiger's back in the top row deviate significantly from the median velocity.

We rank all candidates using (\ref{eq:score_function})
and retain the top $\theta_{P} \%$ candidates as PoTs $\mathcal{P}_{f}$ for this frame.
We found this approach to work quite well in practice.
A few examples  of the top ranking candidates are shown in Fig.~\ref{fig:PoTextraction}. In practice, we use the PoTs shown in Fig.~\ref{fig:PoTextraction}(e,f).

\section{Motion pattern discovery}
\label{sec:discovery}
The input to our motion discovery system is a set of videos $\mathcal{V}$ containing objects of the same class, such as tigers.
The desired output is a set
of clusters $\mathcal{C}=(c_{1},...,c_{k})$ corresponding to motion patterns. 
Each cluster should contain temporal intervals showing the same motion pattern (an interval is any
subsequence of frames).
For the ``tiger'' class, we would like a cluster with tigers
walking, one with tigers turning their head, and so on.
The videos we use (Sec~\ref{sec:eval_protocol}) typically contain several instances of 
different motion patterns each.
For our purposes, it is easier to cluster intervals that correspond
to just one instance of a motion pattern,
and ideally cover the whole duration of that instance.
Hence, we first temporally partition videos into intervals
corresponding to a single motion pattern (Sec.~\ref{sec:partitioning}).
Then we cluster these intervals to discover motion patterns (Sec.~\ref{sec:clustering}).

\subsection{Temporal partitioning}
\label{sec:partitioning}
We first partition videos into shots by thresholding color histogram
differences in consecutive frames~\cite{kim09isce}.
A shot will typically contain several different motion patterns.
For example, a cat may walk for a while, then sit down and finally stretch.
Here, we want to partition the shot into \emph{single-pattern intervals}, \ie, a ``walking'', a ``sitting down'' and a ``stretching'' interval.
Unlike shots, boundaries between such intervals 
cannot be detected using simple color histogram differences.
Instead we partition using two different motion cues:
pauses and periodicity, which we discuss next.

\vspace{-10pt}
\paragraph{Motion-based partitioning.}
We first note that the object often stays still for a brief moment between 
two different motion patterns.
We detect such pauses as sequences of three or more frames without articulated
object motion.  However, some
sequences lack pauses between different related behaviors (\eg, a tiger walking
begins to run). Thus, we also partition based on detected
periodic motion.

\vspace{-10pt}
\paragraph{Periodic motion detector.}
We use time-frequency analysis to detect periodic motion.
We assume periodic motion patterns like
walking, running, or licking
%\footnote{Examples
%of detected periodic motions are available on our
%website~\cite{delpero15cvpr-potswebpage} } 
generate peaks in the frequency domain (examples
are available on our website~\cite{delpero15cvpr-potswebpage}). 
Specifically, we model an input interval as a time sequence 
$s(t)=b_{f^{t}}^{P}$, where $b_{f^{t}}^{P}$ is
the bag-of-words (BoW) of PoTs at frame $f^{t}$. 
We convert $s(t)$ to $C$ one-dimensional sequences (one per codeword) 
and sum the FFTs of the individual sequences in the frequency domain.
If the height of the highest peak is $\geq \theta_{H}$, 
we consider the interval as periodic.
We ensure that the total energy in the frequency domain 
integrates to $1$.
Using the sum of the FFTs makes the approach more
robust, since peaks arise only if several codewords recur with the
same frequency. 

Naively doing time-frequency analysis on an entire interval
typically fails because it might contain both periodic
and non-periodic motion (\eg, a tiger walks for a while
and then sits down). Hence, we consider all possible sub-intervals using
a temporal sliding window and label the one with the highest 
peak as periodic, provided its height $\geq \theta_{H}$. The remaining segments are reprocessed to extract motion patterns with different periods (\eg, walking versus running) until no significant peaks remain. 
For robustness, we only consider sub-intervals
where the period is at least five frames
and the frequency at least three
(\ie, the period repeats at least three times).
We empirically set $\theta_{H} = 0.1$, which
%misses some periodic motion, but
produces very few false positives.

\subsection{Clustering intervals}
\label{sec:clustering}

\paragraph{Interval representation.}
We use $k$-means to form a codebook from a million PoT descriptors randomly sampled from all intervals.
% VF: of course those in I'', or whatever, it really does not matter
We run $k$-means eight times and choose the clustering
with lowest energy to reduce the effects of random 
initialization~\cite{wang_ICCV_2013}. % VF: cuttable
We then represent an interval as a BoW histogram of the PoTs it contains (L1-normalized).

\vspace{-10pt}
\paragraph{Hierarchical clustering.}
We cluster the intervals using hierarchical clustering with complete-linkage~\cite{johnson67psychometrica}.
We found this to perform better than other clustering methods (\eg, single-linkage, $k$-means)
for all the descriptors tested.
As an additional advantage, hierarchical clustering enables one to experiment with different numbers of clusters without re-running the algorithm. % VF: on the edge of not being obvious

\vspace{-10pt}
\paragraph{Distance function.}
Hierarchical clustering requires computing the distance between
pairs of input items. Given BoWs of PoTs $b_{u}$ and $b_{v}$ for intervals $I_{u}$ and $I_{v}$, 
we use
\begin{equation}
\mathrm{d}(I_{u},I_{v})= -\mathrm{exp}\left(~-(1-\mathrm{HI}(b_{u},b_{v})~)~\right) ,
\end{equation}
where $\mathrm{HI}$ denotes histogram intersection. We found this
to perform slightly better than the $\chi^2$ distance for all descriptors tested.
Note that this function can be also used on BoWs of descriptors
other than PoTs.
Additionally, it can be extended to handle different descriptors that use multiple
feature channels, such as Improved DTFs~\cite{wang_ICCV_2013},
which we compare against in the experiments.
In this case, the interval representation is a set of BoWs $(b_{u}^{1},...,b_{u}^{C})$, 
one for each of the $C$ channels.
Following~\cite{wang_ICCV_2013},
we combine all channels into a single distance function
\begin{equation}
\mathrm{d}(I_{u},I_{v})= -\mathrm{exp}\left(-\sum_{i=1}^{C} \frac{1-\mathrm{HI}(b_{u}^{i},b_{v}^{i})}{A_{i}}\right) ,
\end{equation}
where $A_{i}$ is the average value of $(1-\mathrm{HI})$ for channel $i$.

\section{Experiments}
\label{sec:experiments}

In this section, we present our experimental results.

\subsection{Evaluation protocol}
\label{sec:eval_protocol}
\paragraph{Datasets.}
We experiment on two different datasets.
First, we use a dataset of tiger videos
collected from National Geographic documentaries.
This dataset contains roughly two hours of high-resolution, professional footage
divided into 500 shots, for a total of 80,000
frames.
Throughout the experiments, we use various
portions of this dataset:
\begin{compactitem}

\item \emph{Tiger\_fg:} A set of 100 shots with accurate foreground masks~\cite{papazoglou13iccv}, selected manually.

\item \emph{Tiger\_val:} Another set of 100 shots where the
segmentation algorithm works well with no overlap
with Tiger\_fg. We use Tiger\_val to set the parameters
of all the methods we test.

\item \emph{Tiger\_all:} All the shots in the dataset.

\end{compactitem}
Second, we use $100$ shots of the dog class of the YouTube-Objects
dataset~\cite{prest12cvpr}, which mostly contains low-resolution
footage filmed by amateurs.

\vspace{-10pt}
\paragraph{Behavior labels.}
We annotated \emph{each frame} in the dataset independently, choosing from
the behavior labels listed in Table~\ref{table:tiger_intervals}. 
When a frame shows multiple behaviors, we chose the one that happens at the larger
scale (\eg, we choose ``walk" over ``turn
head" and ``turn head" over ``blink").
%If several tigers are visible, we annotated
%the motion primitive of the tiger closest to the camera.
%We annotated $500$ shots from the tiger dataset, and $100$
%from the dog dataset.
As animals move over time, a shot often contains more than one label.
All the labels will be released on our website.

\vspace{-10pt}
\paragraph{Evaluation criteria.}
We use two criteria commonly used for evaluating
clustering methods: \emph{purity} and \emph{Adjusted Rand Index} (ARI)~\cite{rand71jasa}.
Purity is the number of items correctly clustered divided by the
total number of items. An item is correctly clustered
if its label coincides with the most frequent label in its cluster.
While purity is easy to interpret, it only penalizes
assigning two items with different labels to the same cluster.
%Hence, purity grows with the number of clusters,
%and a trivial clustering where each item has its own cluster
%achieves maximum purity.
The ARI instead also penalizes
putting two items with the same label in different clusters.
Further, it is adjusted such that a random clustering
will score close to $0$. 
It is considered a better way to evaluate clustering methods
by the statistics community~\cite{Lawrence_JOC_1985,santos09icann}.
%Another difference is that purity changes
%smoothly as the number of cluster changes, while ARI does not.
%Consider for example splitting a big uniform cluster into two:
%purity would almost not change, while ARI would drop significantly.
%For completeness, we use both criteria in all the experiments.

\vspace{-10pt}
\paragraph{Baseline.}
We compare PoTs to the state-of-the-art Improved Dense Trajectory Features (IDTFs)~\cite{wang_ICCV_2013}. IDTFs combine four different feature
channels aligned with dense trajectories:
Trajectory shape (TS), Histogram of Oriented Gradients (HOG),
Histogram of Optical Flow (HOF), and Motion Boundary Histogram (MBH).
TS is the channel most related to PoTs, as it
encodes the displacement of an individual trajectory
across consecutive frames.
HOG is the only component based on appearance and not on motion.
We also compare against a version of IDTFs where only
trajectories on the foreground segmentation are used. We
call this method fg-IDTFs. We use the same
point tracker~\cite{wang_ICCV_2013} to extract both
IDTFs and PoTs. For PoTs, we do not remove trajectories that are static or are caused by the motion
of the camera. Removing these trajectories improves the performances
of IDTFs~\cite{wang_ICCV_2013}, but in our case they are useful as potential anchors.

\begin{figure}[t]
\begin{center}
\setlength{\tabcolsep}{0.8pt}
\begin{tabular}{c c}
\includegraphics[scale =0.19]{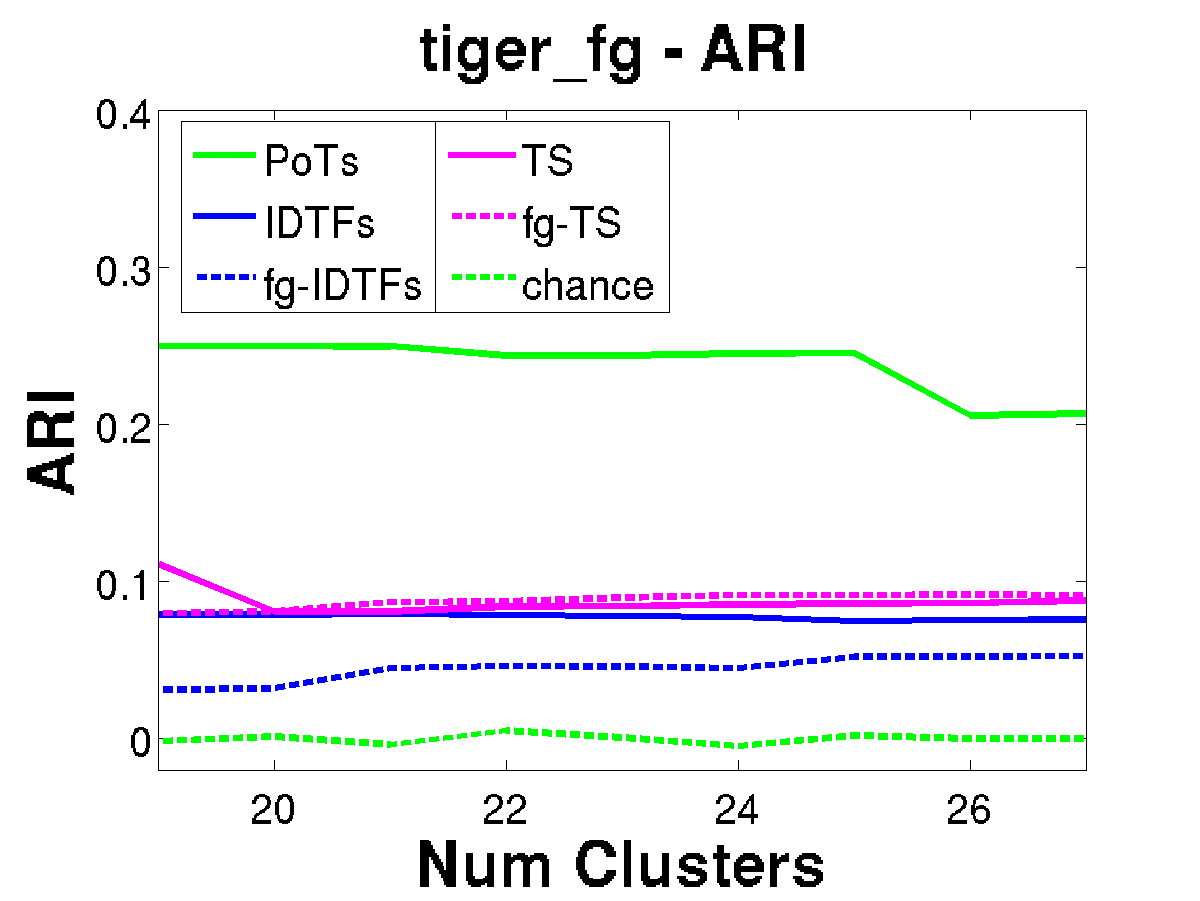} &
\hspace{-10pt}
\includegraphics[scale =0.19]{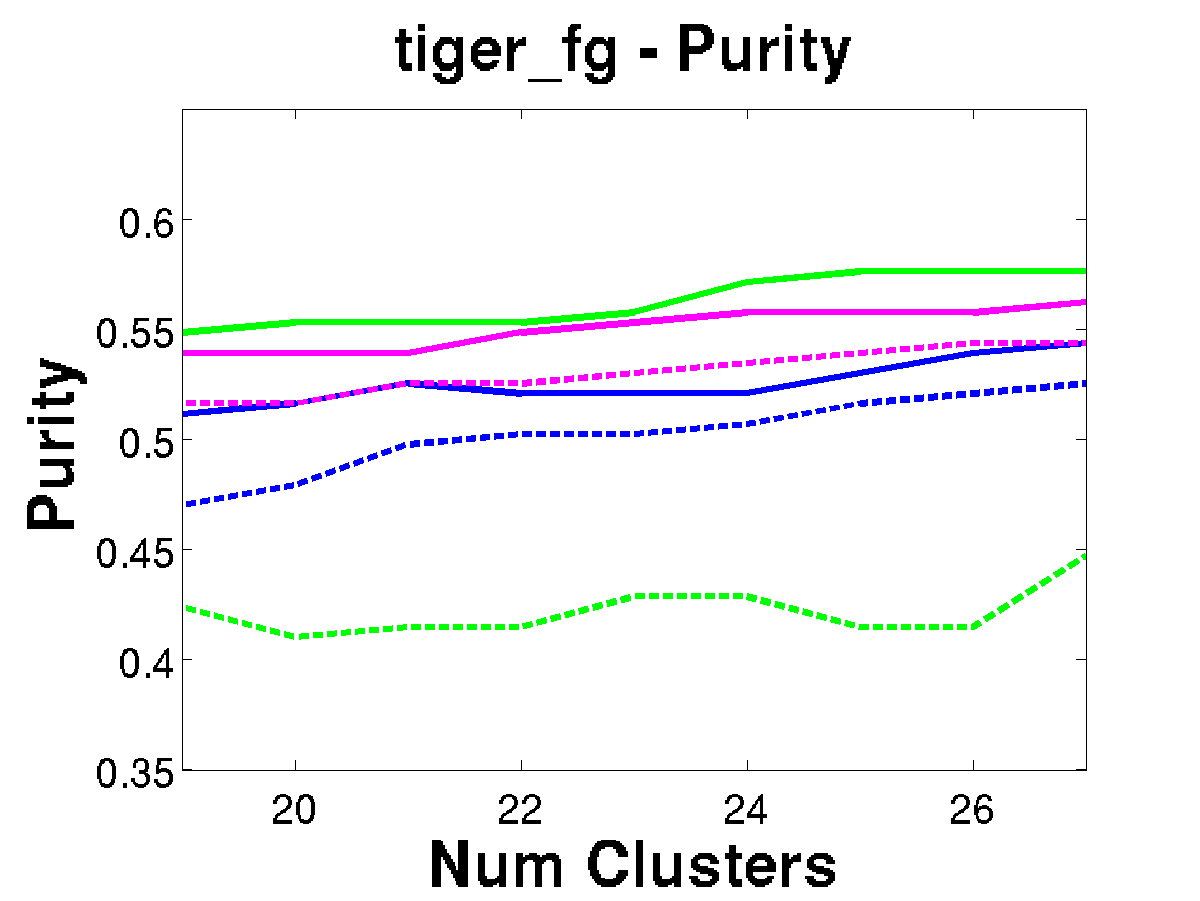} \\
\hline
\includegraphics[scale =0.19]{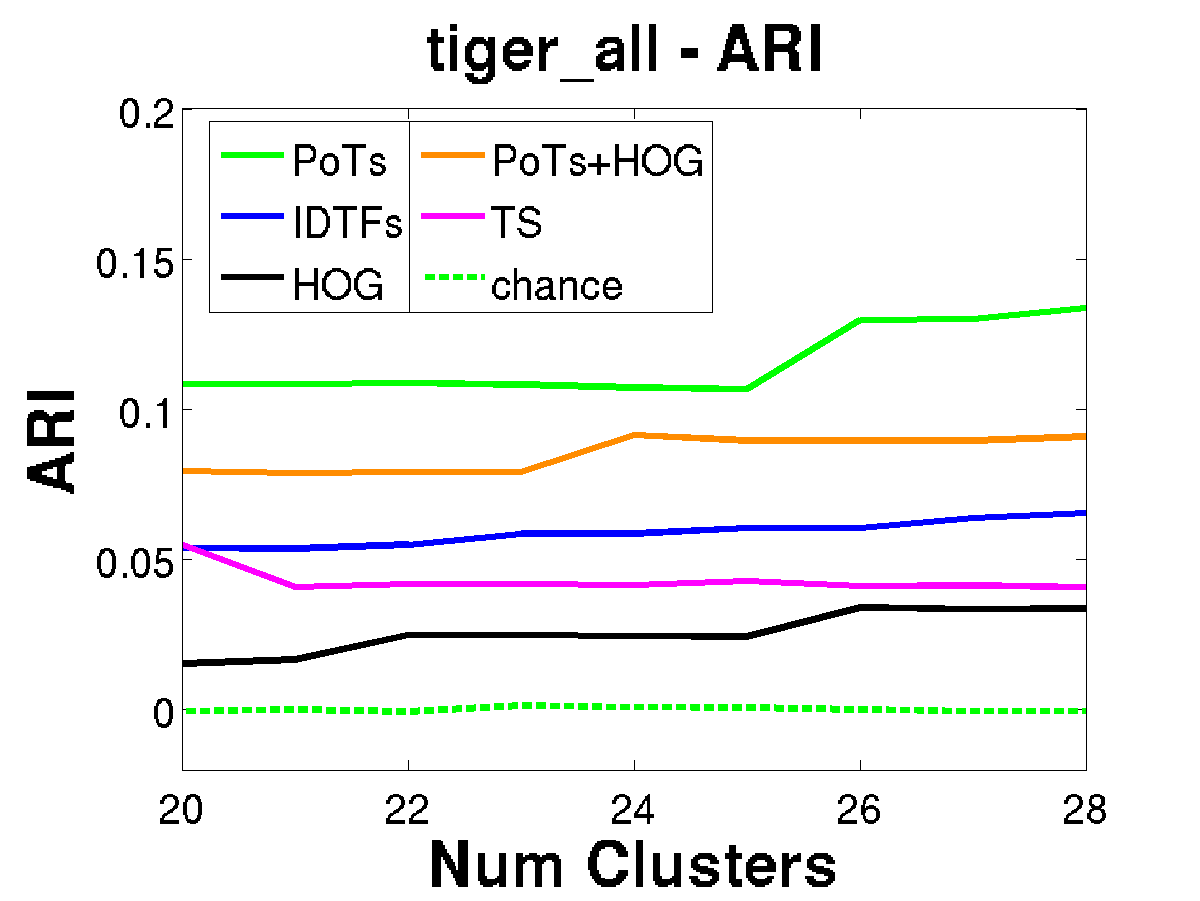}&
\hspace{-10pt}
\includegraphics[scale =0.19]{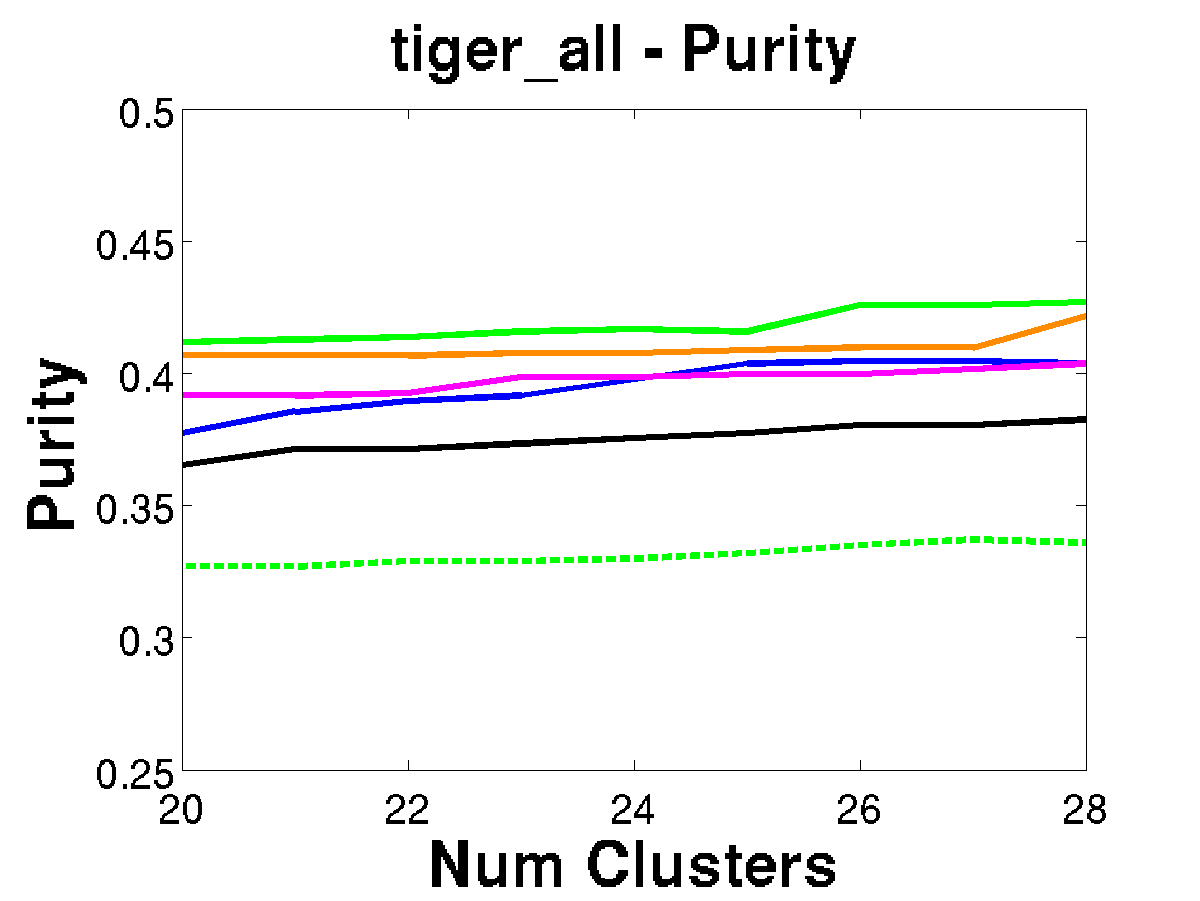} \\
\hline
\includegraphics[scale =0.19]{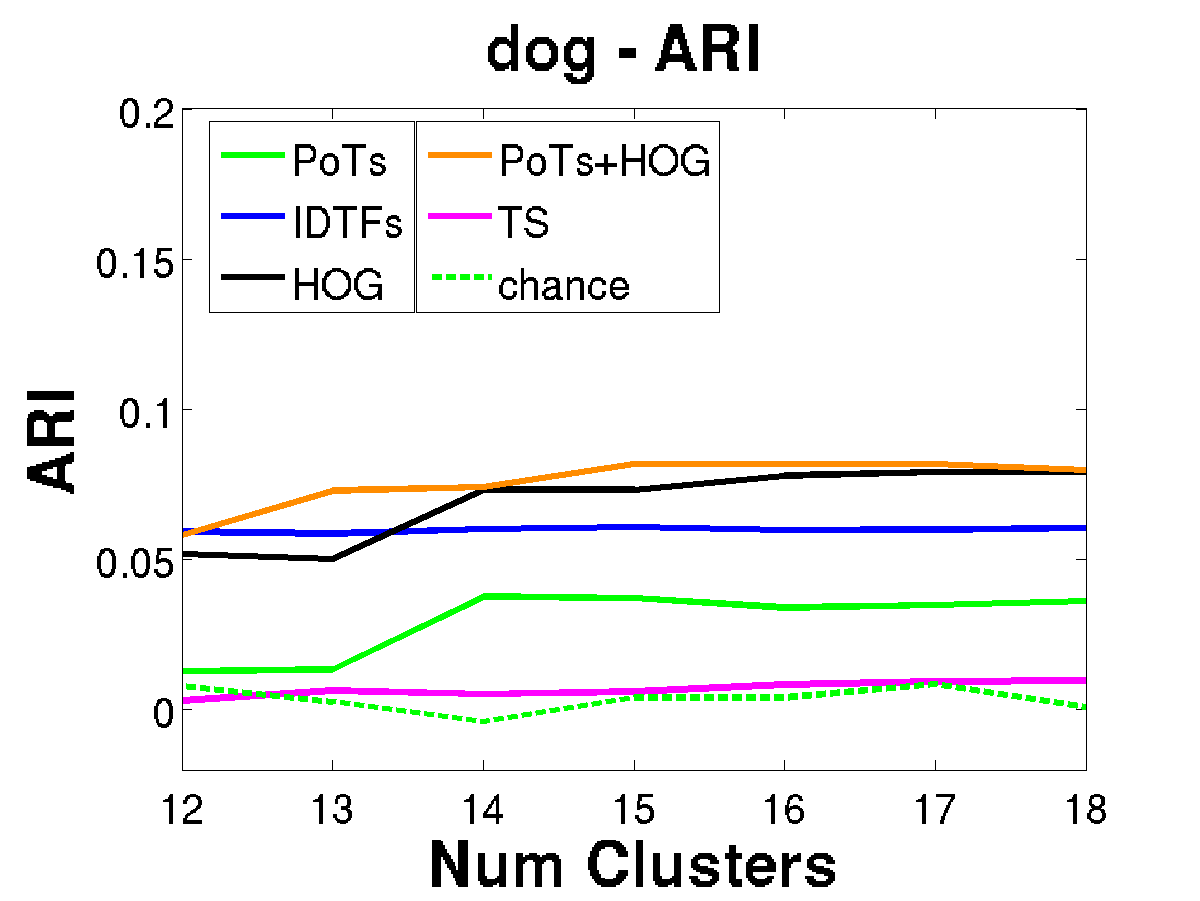}&
\includegraphics[scale =0.19]{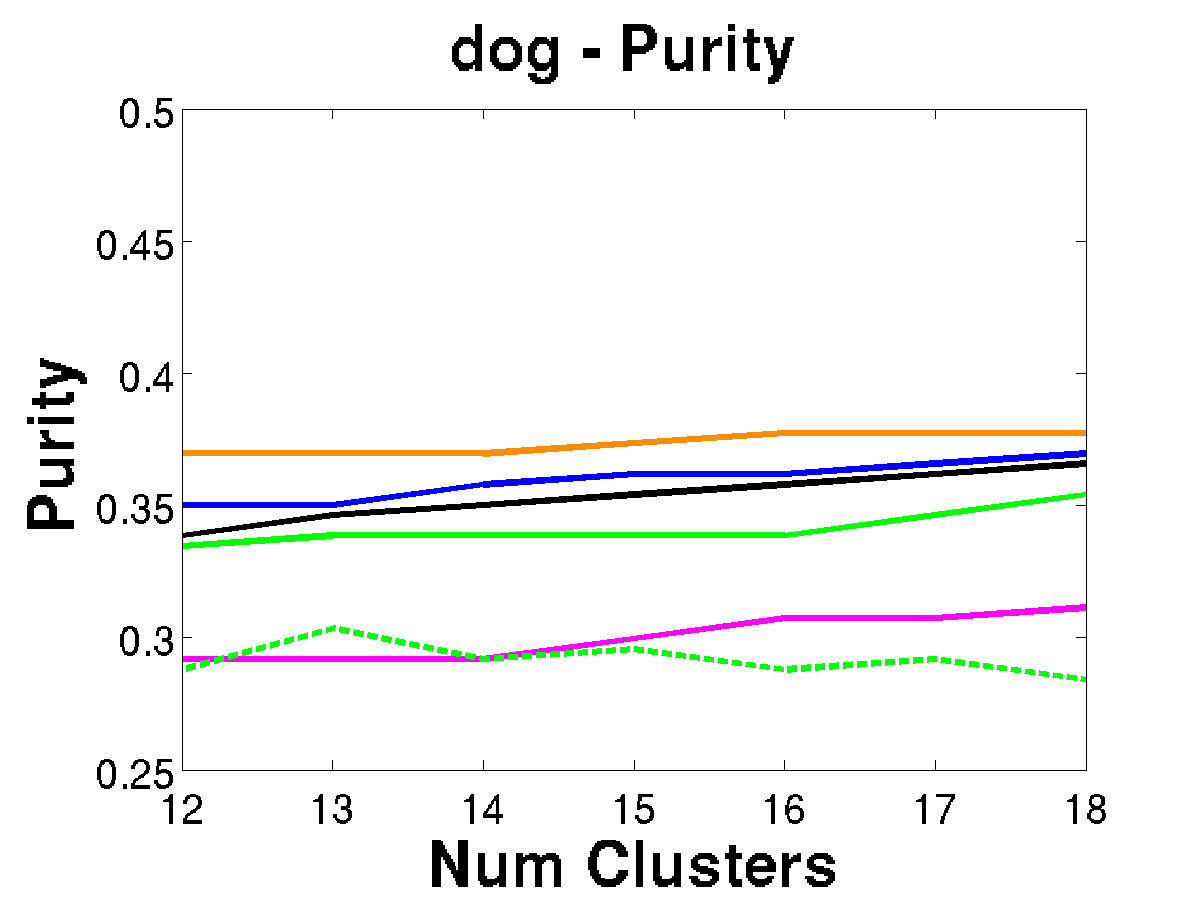} \\
\hline
\includegraphics[scale =0.2]{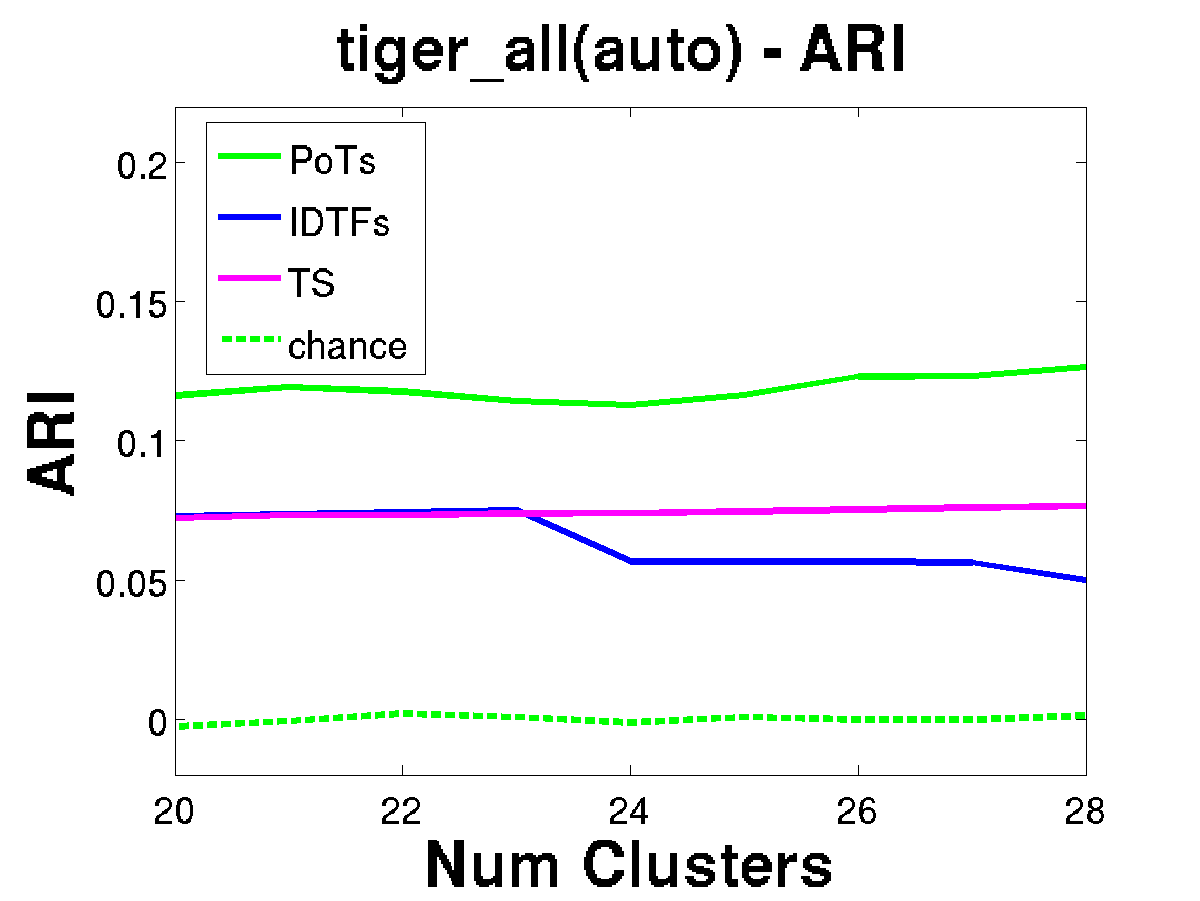} &
\hspace{-10pt}
\includegraphics[scale =0.2]{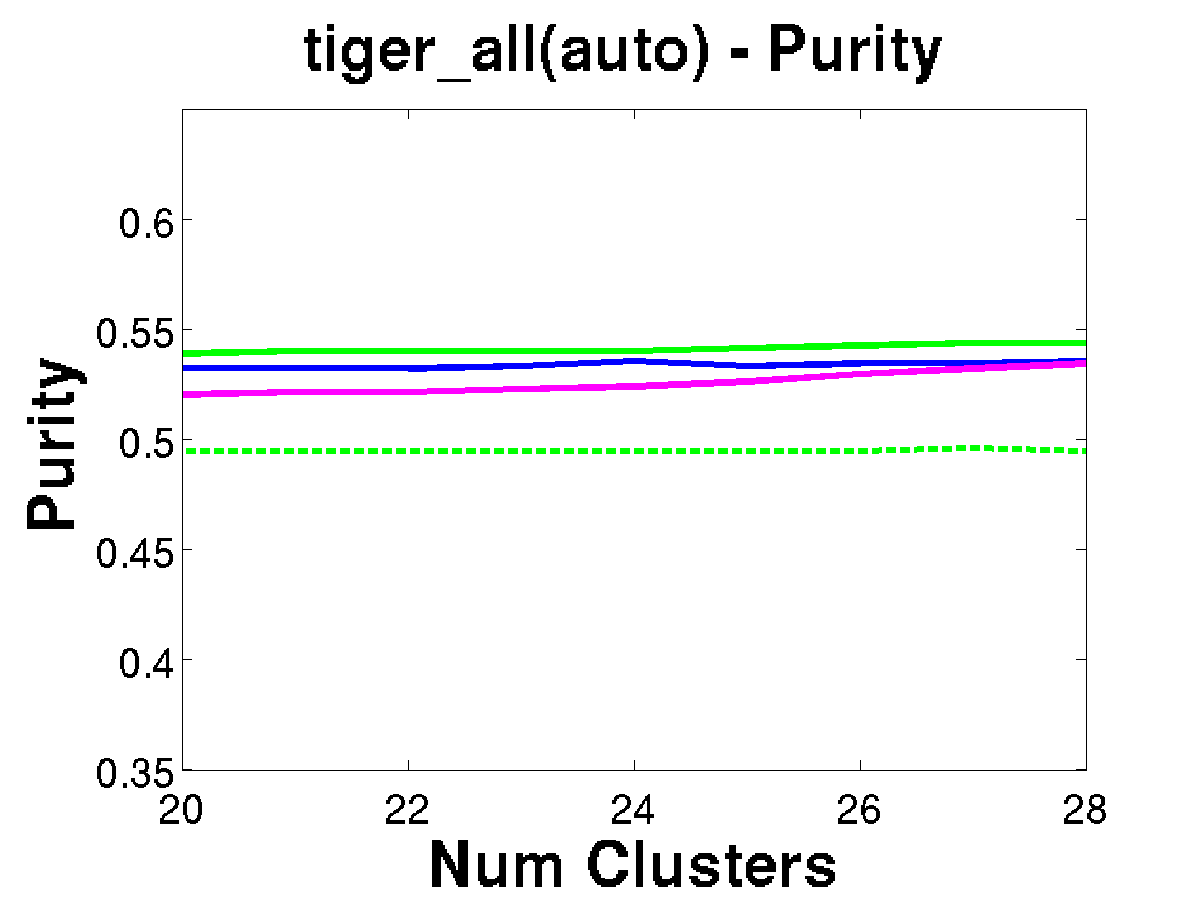}\
\end{tabular}
\end{center}
\vspace{-5pt}
 \caption{\small{
Results of clustering intervals using different descriptors,
evaluated on Adjusted Rand Index (ARI) and purity (see text). 
PoTs result in better clusters than the full 
IDTFs~\cite{wang_ICCV_2013} on tigers (top two rows). Restricting IDTFs to 
the foreground segmentation decreases the performance
on tiger\_fg, where we ensured the segmentation is accurate (top row).
Adding appearance features (PoTs+HOG) is detrimental for tigers
(second row), but improves performance on dogs (third row).
IDTFs perform well for dogs, primarily due to the
contribution of the HOG channel alone (compare the full descriptor, blue, with
the HOG channel only, black, and trajectory shape (TS) channel only, magenta). For
both tigers and dogs, PoTs+HOG performs better than IDTFs.
PoTs also generate higher-quality clusters than the other methods
when we cluster automatically partitioned intervals (bottom row).
}
}
\vspace{-17pt}
\label{fig:resultsmanual}
\vspace{-13pt}
\end{figure}

\begin{table*}[t]
\begin{center}
\scriptsize
\setlength{\tabcolsep}{2.8pt}
%\begin{tabular}{|c||c|c|c|c|c|c|c|c|c|c|c|c|c|c|c|c|c|c|c|c|c|c|c|}
\begin{tabular}{lccccccccccccccccccccccc}
\toprule
%\hline
\textbf{} &\textbf{walk} &\textbf{turn}  &\textbf{sit} &\textbf{tilt} &\textbf{stand} &\textbf{drag} &\textbf{wag} &\textbf{walk} &\textbf{run} &\textbf{turn} &\textbf{jump} &\textbf{raise} &\textbf{open} &\textbf{close} &\textbf{blink} &\textbf{slide} &\textbf{drink} &\textbf{chew} &\textbf{lick} &\textbf{climb} &\textbf{roll} &\textbf{scratch} &\textbf{swim} \tabularnewline
\textbf{Partitions} &\textbf{} &\textbf{head}  &\textbf{down} &\textbf{head} &\textbf{up} &\textbf{} &\textbf{tail} &\textbf{back} &\textbf{} &\textbf{} &\textbf{} &\textbf{paw} &\textbf{mouth} &\textbf{mouth} &\textbf{} &\textbf{leg} &\textbf{} &\textbf{} &\textbf{} &\textbf{} &\textbf{} &\textbf{} &\textbf{} \tabularnewline
%\hline
\midrule
\textbf{\tiny whole shots} & 259 & 24 & 5 & 17 & 5 & 4 & 1 & 5 & 16 & 3 & 9 & 0 & 4 & 0 & 0 & 1 & 7 & 4 & 6 & 1 & 3 & 1 & 3 \tabularnewline
%\hline
\textbf{\tiny pause} & 272 & 76 & 11 & 30 & 9 & 4 & 2 & 6 & 16 & 2 & 11 & 2 & 11 & 3 & 10 & 2 & 7 & 5 & 7 & 1 & 5 & 1 & 3 \tabularnewline
%\hline
\textbf{\tiny pause+periods} & \textbf{273} & \textbf{80} & \textbf{13} & \textbf{33} & \textbf{9} & \textbf{4} & \textbf{2} & \textbf{6} & \textbf{16} & \textbf{3} & \textbf{11} & \textbf{2} & \textbf{12} & \textbf{3} & \textbf{11} & \textbf{3} & \textbf{8} & \textbf{5} & \textbf{7} & \textbf{1} & \textbf{5} & \textbf{1} & \textbf{3} \tabularnewline
%\hline
\textbf{\tiny ground truth} & 289 & 148 & 27 & 77 & 24 & 4 & 4 & 10 & 23 & 18 & 20 & 6 & 40 & 28 & 39 & 13 & 12 & 7 & 19 & 1 & 5 & 2 & 3 \tabularnewline
%\hline
\bottomrule
\end{tabular}
\begin{tabular}{lcccccccccccccccc}
\toprule
%\hline
\textbf{} &\textbf{walk} &\textbf{turn}  &\textbf{sit} &\textbf{tilt} &\textbf{stand} &\textbf{walk} &\textbf{run} &\textbf{turn} &\textbf{jump} &\textbf{open} &\textbf{close} &\textbf{blink} &\textbf{slide} &\textbf{lick} &\textbf{push} \tabularnewline
\textbf{Partitions} &\textbf{} &\textbf{head}  &\textbf{down} &\textbf{head} &\textbf{up} &\textbf{back} &\textbf{} &\textbf{} &\textbf{} &\textbf{mouth} &\textbf{mouth} &\textbf{} &\textbf{leg} &\textbf{} &\textbf{skateboard} \tabularnewline
%\hline
\midrule
\textbf{\tiny whole shots} & 25 & 4 & 0 & 1 & 0 & 0 & 10 & 0 & 4 & 0 & 0 & 0 & 0 & 0 & 13  \tabularnewline
%\hline
\textbf{\tiny pause} & 29 & 5 & 0 & 1 & 0 & 0 & 12 & 2 & 5 & 0  & 0 & 0  & 0 & 0 & 16  \tabularnewline
%\hline
\textbf{\tiny pause+periods} & \textbf{29} & \textbf{9} & \textbf{0}  & \textbf{3} & \textbf{0} & \textbf{1} & \textbf{13} & \textbf{5} & \textbf{5} & \textbf{0} & \textbf{0} & \textbf{0} & \textbf{0} & \textbf{0} & \textbf{16} \tabularnewline
%\hline
\textbf{\tiny ground truth} & 39 & 25 & 1 & 12 & 1 & 2 & 20 & 14 & 8 & 2 & 1 & 2 & 1 & 1 & 19  \tabularnewline
%\hline
\bottomrule
\end{tabular}
 \vspace{-8pt}
\end{center}
 \caption{\small
Number of intervals recovered per behavior on tigers (top) and dogs (bottom). Pause+periods consistently dominates others.
 \label{table:tiger_intervals}}
 \vspace{-5pt}
\end{table*}
 \vspace{-8pt}

\vspace{-10pt}
\paragraph{Calibration.}
We use Tiger\_val to set the PoT selection threshold $\theta_{P}$
(Sec.~\ref{sec:extraction}) and the PoT codebook
size $K$ (sec.~\ref{sec:clustering}) using coarse grid search.
As objective function, we used the ARI achieved by our 
method with the number of clusters equal to the true number of behaviors.
%We used interval $[0.05,0.35]$ with a step of $0.05$ for
%$\theta_{P}$,  and $[800,6000]$ with a step
%of $800$ for $C$. 
The chosen parameters are $\theta_{P}=0.15$ and $K=800$.
We tuned the IDTF codebook size similarly;
the best size was $4000$. Interestingly, this is the same value as chosen by Wang et al.~\cite{wang_ICCV_2013}
on completely different data.

\subsection{Evaluating PoTs}
\label{sec:evalpots}
We first evaluate PoTs in a simplified scenario
where the correct single-pattern partitioning is given,
\ie, we partition shots using frames where the ground-truth label changes as boundaries. 
This allows us to evaluate the PoT representation separately 
from our method for automatic interval discovery (Sec.~\ref{sec:partitioning}).
%The number of manual intervals 
%grouped by motion primitive is shown in the bottom row of Table~\ref{table:tiger_intervals}.
We compare clustering using BoWs of PoTs to 
clustering using BoWs of IDTFs in Fig.~\ref{fig:resultsmanual}.
As the true number of clusters
is usually not known a priori, each plot
shows performance as a function of the number of clusters. 
The mid value on the horizontal axis corresponds
to the true number of clusters ($23$ for tigers, $15$ for dogs).

\vspace{-12pt}
\paragraph{Evaluation on tigers.} 
The clusters found using PoTs are better in both purity and ARI (Fig.~\ref{fig:resultsmanual}).
The gain over IDTFs is
larger on Tiger\_fg (top row), where PoTs benefit from the accurate
estimate of the foreground. Here, PoTs also outperform fg-IDTFs. 
This shows that the power of our representation resides in the principled
use of pairs, not just in exploiting the foreground segmentation
to remove background trajectories.
Results on Tiger\_all (second row) show that PoTs can also cope with imperfect segmentation.
%We also compare against individual channels of the IDTFs (Fig~\ref{fig:resultsmanual}).
%HOF and MBH perform very poorly and are not shown here.

Consider now the individual IDTFs channels.
HOG performs poorly and causes the complete IDTFs to perform worse than their TS channel alone, although both are inferior to PoTs.
Similarly, adding the HOG channel to PoTs performs
worse than pure motion PoTs but is still better than
IDTFs.
Appearance is in general not suitable for discovering fine-grained motion patterns.
It is particularly misleading in a class like ``tiger" where different instances have similar color and texture.
The HOF and MBH channels of IDTF perform poorly on their own and are not shown here.

\vspace{-15pt}
\paragraph{Evaluation on dogs.}
The complete IDTF descriptors perform better than PoTs on the dog dataset 
(Fig.~\ref{fig:resultsmanual}, third row). However,
the HOG channel is doing most of the work in this case.
The dog shots come from only eight different videos, each showing
one particular dog performing $1$--$2$ behaviors in the same scene.
Hence, HOG performs well by trivially clustering together
intervals from the same video. 
If we equip PoTs with the HOG channel, they outperform
the complete IDTFs. Similarly, when considering trajectory motion alone,
PoTs outperform the IDTF TS channel. These experiments confirm that PoTs are a better representation for articulated objects than IDTF
also on the dog data.

\vspace{-8pt}
\luca{Added this paragraph, check}
\paragraph{Comparison to motion primitives~\cite{Yang_PAMI_2013}.}
Last, we compare to the method of~\cite{Yang_PAMI_2013}, which is based on motion primitives.
We compare on the KTH dataset~\cite{KTH} in their setting. 
It contains $100$ shots
for each of six different human actions (\eg walking, hand clapping).
%all manually trimmed to the duration of the action.
As before, we cluster all shots using the PoT representation: for the true number 
of clusters (6), we achieve 59\% purity, compared to their 38\% (Fig. 9 in~\cite{Yang_PAMI_2013}).
For this experiment, we incorporated an R-CNN person detector~\cite{girshick14cvpr}
%(trained on ImageNet)
into the foreground segmentation algorithm~\cite{papazoglou13iccv}
to better segment the actors.
\begin{table}
 \vspace{-5pt}
\begin{center}
\small
\setlength{\tabcolsep}{1.8pt}
\begin{tabular}{lcccc}
%\begin{tabular}{|c|c|c|c|c|}
%\hline
\toprule
&\textbf{whole shots} & \textbf{pauses} & \textbf{pauses+periods} & \textbf{ground truth} \tabularnewline
%\hline
\midrule
tiger \# intervals &480 & 719 & \textbf{885} & 1026 \tabularnewline
tiger uniformity &0.78 & 0.85 & \textbf{0.87} & 1 \tabularnewline
%\hline
\hline
%\hline
\midrule
dog \# intervals & 80 & 115 & \textbf{219} & 260 \tabularnewline
dog uniformity &0.72 & 0.80 & \textbf{0.88} & 1 \tabularnewline
\bottomrule
\end{tabular}
\end{center}
\vspace{-4pt}
 \caption{\small
Interval uniformity for different partitioning methods.
%on tiger\_all and dog.
%Partitioning shots based on pauses and periodicity is best and approaches the groundtruth partitioning.
Pauses+periods consistently outperforms alternatives.
 \label{table:interval_uniformity}}
\vspace{-12pt}
\end{table}
\vspace{-1pt}

\begin{figure*}[t]
\begin{center}
\includegraphics[scale =0.16]{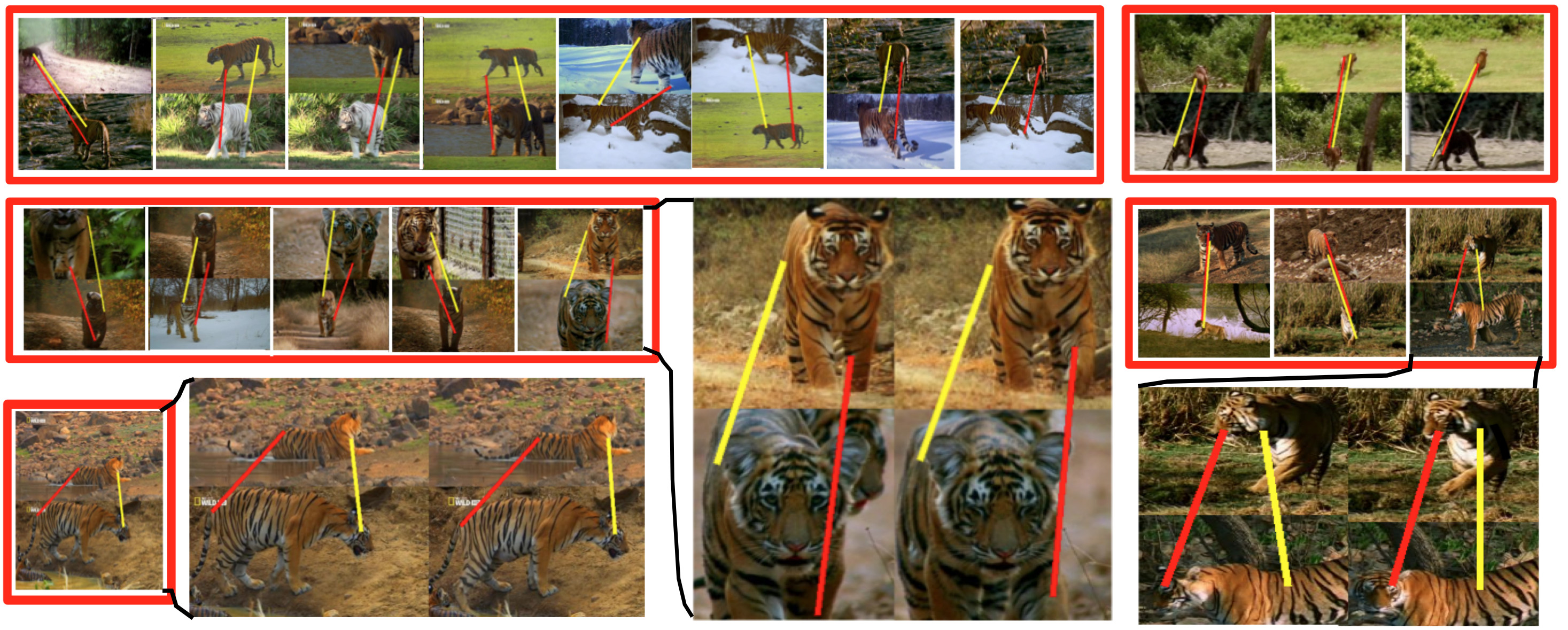}
\end{center}
\vspace{-15pt}
 \caption{\small{
Behaviors discovered by clustering consistent motion patterns.
Each red rectangle displays a few pairs of intervals
from one cluster, on which we connect the anchors (yellow)
and swings (red) of two individual PoTs that are close in descriptor space.
The enlarged version show how the connected PoTs evolve through time
and give a snapshot of the captured motion pattern in each cluster.
%For example, the rotation of the head wrt other part
%of the body in the ``turning head" cluster" (bottom right).
The behaviors shown are: two different ways of walking (left, top and middle), sitting
(bottom left), running (top right), and turning head (bottom right).
 }
 }
\label{fig:qualitative}
\vspace{-10pt}
\end{figure*}

\subsection{Evaluating motion discovery}
\label{sec:evalclustering}
%\paragraph{Evaluation of partitioning.} 
We now evaluate our method for partitioning into single-pattern intervals.
Let the \emph{interval uniformity} be the number of frames with the
most frequent label in the interval, divided by the total number of frames.
Our baseline is the average interval uniformity
of the original shots without any partitioning.
The combination of pauses and  periodicity partitioning improves the 
average interval uniformity (Table~\ref{table:interval_uniformity}).
This is very promising, since the average interval uniformity is close to $90\%$,
and the number of intervals found approaches the ground-truth number. 
In Table~\ref{table:tiger_intervals} we report the number of single-pattern 
intervals found by each method, grouped by motion pattern. Here,
we only increase the count for intervals from distinct shots.
%Here, all intervals with the same label that come from the same shot count as just one.
Otherwise, we could approach ground truth by simply chopping one continuous behavior into smaller and smaller pieces. We chose this counting method 
because finding instances of the same pattern performed by different tigers is
our goal.
If we were to cluster whole shots, many patterns
would be lost, and only a few dominant classes would emerge 
from the data. Instead, our method finds intervals
for each label.

%\vspace{-10pt}
%\paragraph{Clustering partitioned intervals.}
We report purity and ARI for the clusters 
of partitioned intervals. 
As the ground-truth label for a partitioned interval, which may not coincide exactly with a ground-truth interval, we use the label of the majority of the frames in the interval.
%For evaluation, we label a partitioned interval with the most frequent frame label.
To make this comparison fair, we evaluated the IDTF descriptors on the
same single-pattern intervals.
As before, PoTs outperform IDTFs (Fig.~\ref{fig:resultsmanual}, bottom row). 
%(which used the PoT
%descriptor to detect periodic motion). We tried partitioning using the IDTF
%descriptor instead, but the resulting partition was much worse.
Finally, we
show a few qualitative examples of the clusters found by our method in
Fig.~\ref{fig:qualitative}.
%\footnote{Example videos for both dogs and tiger clusters can be found in the supplemental material.}

\vspace{-5pt}
\section{Discussion}
\vspace{-4pt}
We emphasize that the only supervision in the entire process is the initial
video label (\ie, we know the video contains a tiger or dog, respectively) and
that the only cue used is motion, encoded by the PoT descriptor.

Appearance features  have proved useful for traditional
action recognition tasks~\cite{UCF101, Sports1M}, since many activities
are strongly characterized by the background and the apparel involved 
(\eg, diving can be recognized from the appearance of swimsuits, or a diving board
with a pool below).
%(\eg, diving can be
%recognized from the appearance of of a diver in a swimsuit, standing on the diving board
%with the pool visible below)
% humans performing a
%particular action often wear apparel specific to that activity and appear
%against background characteristic for that activity . 
The dog dataset fits this paradigm: the
appearance of the individual dog and the background was tightly correlated with
the dog's behavior (\eg, only one dog knew how to skateboard) and so adding
appearance should be beneficial. Because PoTs and appearance features are
complementary, we see the expected performance boost by adding the additional
information. However, the tigers dataset shows that adding appearance features can be detrimental. 
Tigers varied in appearance (orange and white
tigers, cubs and adults, etc.) but all tigers performed a variety of behaviors.
On this dataset, the motion-only PoT descriptor outperforms all tested
alternatives that included appearance information.

%The essential feature of PoTs is the use of trajectory
%pairs, so that a collection of PoTs can encode detailed information about the 
%relative motion between many different parts of an object.
An essential feature of our method is that a collection of PoTs 
can encode detailed information about the 
relative motion between many different parts of an object.
PoT anchors are scattered across the
object; each may move with its own unique trajectory.
Simplifying PoTs to a star-like model 
where all anchors coincide with the center of mass of the
object (\ie, normalizing by the dominant object motion) would
result in a loss of expressive power and would be less robust for highly
deformable objects.

PoTs are selected bottom-up and need not relate to the kinematic structure of
an object class.
This allows the extraction process to apply to any object and to leverage those PoTs 
that are discriminative for the particular class rather than being limited to pre-defined relationships
We have shown that clustering built on top of PoTs
finds motion patterns that are consistent across many
shots. While many common behaviors (\eg, walking) are cyclic, our method
focuses instead on consistency across occurrences 
rather than periodicity within an occurrence.
%, enabling us to discover behaviors such as a tiger turning its head.
Periodic motion is exploited during partitioning, but the clustering procedure
itself makes no such assumption, enabling us to discover behaviors such as a tiger turning its head.

\vspace{-8pt}
\paragraph{Acknowledgments.}
\luca{Added. Note they do NOT have to be in the first 8 pages}
We are very grateful to Anestis Papazoglou for help with the data collection,
and to Shumeet Baluja for his helpful comments. This work was partly funded by a Google Faculty Research Award,
and by ERC Starting Grant ``Visual Culture for Image Understanding''.

\goodbreak
{\small
\bibliographystyle{ieee}
\bibliography{../../bibtex/shortstrings,../../bibtex/calvin,../../bibtex/vggroup}

\begin{thebibliography}{10}\itemsep=-1pt

\bibitem{BourdevMalikICCV09}
L.~Bourdev and J.~Malik.
\newblock Poselets: Body part detectors trained using 3d human pose
  annotations.
\newblock In {\em ICCV}, 2009.

\bibitem{Cootes1998ECCV}
T.~Cootes, G.~Edwards, and C.~Taylor.
\newblock Active appearance models.
\newblock In {\em ECCV}, 1998.

\bibitem{delpero15cvpr-potswebpage}
L.~Del~Pero, S.~Ricco, R.~Sukthankar, and V.~Ferrari.
\newblock Dataset for articulated motion discovery using pairs of trajectories.
\newblock \url{http://groups.inf.ed.ac.uk/calvin/proj-pots/page/}, 2015.

\bibitem{Felzenszwalb03pictorialstructures}
P.~F. Felzenszwalb and D.~P. Huttenlocher.
\newblock Pictorial structures for object recognition.
\newblock {\em IJCV}, 61(1):55--79, 2005.

\bibitem{girshick14cvpr}
R.~Girshick, J.~Donahue, T.~Darrell, and J.~Malik.
\newblock Rich feature hierarchies for accurate object detection and semantic
  segmentation.
\newblock In {\em CVPR}, 2014.

\bibitem{WeizmannActions}
L.~Gorelick, M.~Blank, E.~Shechtman, M.~Irani, and R.~Basri.
\newblock Actions as space-time shapes.
\newblock {\em IEEE Trans. on PAMI}, 29(12):2247--2253, December 2007.

\bibitem{HospedalesICCV09}
T.~Hospedales, S.~Gong, and T.~Xiang.
\newblock A {M}arkov clustering topic model for mining behaviour in video.
\newblock In {\em ICCV}, 2009.

\bibitem{Hu2006PAMI}
W.~Hu, X.~Xiao, Z.~Fu, D.~Xie, T.~Tan, and S.~Maybank.
\newblock A system for learning statistical motion patterns.
\newblock {\em IEEE Trans. on PAMI}, 28(9):1450--1464, 2006.

\bibitem{Lawrence_JOC_1985}
L.~Hubert and P.~Arabie.
\newblock Comparing partitions.
\newblock {\em Journal of Classification}, 2(1):193--218, 1985.

\bibitem{Jain2013}
A.~Jain, A.~Gupta, M.~Rodriguez, and L.~Davis.
\newblock Representing videos using mid-level discriminative patches.
\newblock In {\em CVPR}, 2013.

\bibitem{Jiang2012}
Y.-G. Jiang, Q.~Dai, X.~Xue, W.~Liu, and C.-W. Ngo.
\newblock Trajectory-based modeling of human actions with motion reference
  points.
\newblock In {\em ECCV}, 2012.

\bibitem{THUMOS2014}
Y.-G. Jiang, J.~Liu, G.~Toderici, A.~R. Zamir, I.~Laptev, M.~Shah, and
  R.~Sukthankar.
\newblock {THUMOS:} {ECCV} workshop on action recognition with a large number
  of classes, 2014.

\bibitem{johnson67psychometrica}
S.~C. Johnson.
\newblock Hierarchical clustering schemes.
\newblock {\em Psychometrika}, 2:241--254, 1967.

\bibitem{Sports1M}
A.~Karpathy, G.~Toderici, S.~Shetty, T.~Leung, R.~Sukthankar, and L.~Fei-Fei.
\newblock Large-scale video classification with convolutional neural networks.
\newblock In {\em CVPR}, 2014.

\bibitem{kim09isce}
W.-H. Kim and J.-N. Kim.
\newblock An adaptive shot change detection algorithm using an average of
  absolute difference histogram within extension sliding window.
\newblock In {\em ISCE}, 2009.

\bibitem{HMDB}
H.~Kuehne, H.~Jhuang, E.~Garrote, T.~Poggio, and T.~Serre.
\newblock {HMDB}: A large video database for human motion recognition.
\newblock In {\em ICCV}, 2011.

\bibitem{HMDB51}
H.~Kuehne, H.~Jhuang, E.~Garrote, T.~Poggio, and T.~Serre.
\newblock Hmdb: a large video database for human motion recognition.
\newblock In {\em ICCV}, 2011.

\bibitem{kuettel10cvpr}
D.~Kuettel, M.~Breitenstein, L.~van Gool, and V.~Ferrari.
\newblock What's going on? {D}iscovering spatio-temporal dependencies in
  dynamic scenes.
\newblock In {\em CVPR}, 2010.

\bibitem{Leordeanu2007}
M.~Leordeanu, M.~Hebert, and R.~Sukthankar.
\newblock Beyond local appearance: Category recognition from pairwise
  interactions of simple features.
\newblock In {\em CVPR}, 2007.

\bibitem{Mahadevan2010}
V.~Mahadevan, W.~Li, V.~Bhalodia, and N.~Vasconcelos.
\newblock Anomaly detection in crowded scenes.
\newblock In {\em CVPR}, 2010.

\bibitem{Matikainen2009}
P.~Matikainen, M.~Hebert, and R.~Sukthankar.
\newblock Trajections: Action recognition through the motion analysis of
  tracked features.
\newblock In {\em ICCV Workshop on Video-Oriented Object and Event
  Classification}, 2009.

\bibitem{Matikainen2010}
P.~Matikainen, M.~Hebert, and R.~Sukthankar.
\newblock Representing pairwise spatial and temporal relations for action
  recognition.
\newblock In {\em ECCV}, 2010.

\bibitem{Messing2009}
R.~Messing, C.~Pal, and H.~Kautz.
\newblock Activity recognition using the velocity histories of tracked
  keypoints.
\newblock In {\em ICCV}, 2009.

\bibitem{narayan14cvpr}
S.~Narayan and K.~R. Ramakrishnan.
\newblock A cause and effect analysis of motion trajectories for modeling
  actions.
\newblock In {\em CVPR}, 2014.

\bibitem{papazoglou13iccv}
A.~Papazoglou and V.~Ferrari.
\newblock Fast object segmentation in unconstrained video.
\newblock In {\em ICCV}, December 2013.

\bibitem{prest12cvpr}
A.~Prest, C.~Leistner, J.~Civera, C.~Schmid, and V.~Ferrari.
\newblock Learning object class detectors from weakly annotated video.
\newblock In {\em CVPR}, 2012.

\bibitem{ramanan06pami}
D.~Ramanan, A.~Forsyth, and K.~Barnard.
\newblock Building models of animals from video.
\newblock {\em IEEE Trans. on PAMI}, 28(8):1319 -- 1334, 2006.

\bibitem{rand71jasa}
W.~M. Rand.
\newblock Objective criteria for the evaluation of clustering methods.
\newblock {\em Journal of the American Statistical Association}, 66:846--850,
  1971.

\bibitem{raptis12cvpr}
M.~Raptis, I.~Kokkinos, and S.~Soatto.
\newblock Discovering discriminative action parts from mid-level video
  representations.
\newblock In {\em CVPR}, 2012.

\bibitem{raptis10eccv}
M.~Raptis and S.~Soatto.
\newblock Tracklet descriptors for action modeling and video analysis.
\newblock In {\em ECCV}, 2010.

\bibitem{UTInteraction}
M.~S. Ryoo and J.~K. Aggarwal.
\newblock Spatio-temporal relationship match: Video structure comparison for
  recognition of complex human activities.
\newblock In {\em ICCV}, 2009.

\bibitem{santos09icann}
J.~M. Santos and M.~Embrechts.
\newblock On the use of the adjusted rand index as a metric for evaluating
  supervised classification.
\newblock In {\em ICANN}, 2009.

\bibitem{KTH}
C.~Schuldt, I.~Laptev, and B.~Caputo.
\newblock Recognizing human actions: A local svm approach.
\newblock In {\em Proc. ICPR}, 2004.

\bibitem{UCF101}
K.~Soomro, A.~R. Zamir, and M.~Shah.
\newblock {UCF101}: A dataset of 101 human action classes from videos in the
  wild.
\newblock Technical Report CRCV-TR-12-01, University of Central Florida, 2012.

\bibitem{Tang2013}
K.~Tang, R.~Sukthankar, J.~Yagnik, and L.~Fei-Fei.
\newblock Discriminative segment annotation in weakly labeled video.
\newblock In {\em CVPR}, 2013.

\bibitem{Turaga2008}
P.~Turaga, R.~Chellappa, V.~Subrahmanian, and O.~Udrea.
\newblock Machine recognition of human activities: A survey.
\newblock {\em IEEE T-CVST}, 18(11):1473--1488, 2008.

\bibitem{Wang_2011_CVPR}
H.~Wang, A.~Kl\"{a}ser, C.~Schmid, and L.~Cheng-Lin.
\newblock {Action Recognition by Dense Trajectories}.
\newblock In {\em CVPR}, 2011.

\bibitem{wang_ICCV_2013}
H.~Wang and C.~Schmid.
\newblock Action recognition with improved trajectories.
\newblock In {\em ICCV}, 2013.

\bibitem{Wang2014}
L.~Wang, Y.~Qiao, and X.~Tang.
\newblock Video action detection with relational dynamic-poselets.
\newblock In {\em ECCV}, 2014.

\bibitem{Wang2009PAMI}
X.~Wang, X.~Ma, and W.~Grimson.
\newblock Unsupervised activity perception in crowded and complicated scenes
  using hierarchical {Bayesian} models.
\newblock {\em IEEE Trans. on PAMI}, 31(3):539--555, 2009.

\bibitem{Weinland2010}
D.~Weinland, R.~Ronfard, and E.~Boyer.
\newblock A survey of vision-based methods for action representation,
  segmentation and recognition.
\newblock {\em CVIU}, 115(2):224--241, 2010.

\bibitem{Yang2010}
S.~Yang, M.~Chen, D.~Pomerleau, and R.~Sukthankar.
\newblock Food recognition using statistics of pairwise local features.
\newblock In {\em CVPR}, 2010.

\bibitem{Yang_PAMI_2013}
Y.~Yang, I.~Saleemi, and M.~Shah.
\newblock Discovering motion primitives for unsupervised grouping and one-shot
  learning of human actions, gestures, and expressions.
\newblock {\em IEEE Trans. on PAMI}, 35(7):1635--1648, 2013.

\bibitem{MSRActions}
J.~Yuan, Z.~Liu, and Y.~Wu.
\newblock Discriminative subvolume search for efficient action detection.
\newblock In {\em CVPR}, 2009.

\bibitem{Zhao2011}
X.~Zhao and G.~Medioni.
\newblock Robust unsupervised motion pattern inference from video and
  applications.
\newblock In {\em ICCV}, 2011.

\end{thebibliography}
}

\end{document}